# Letter-level Online Writer Identification

Zelin Chen, Hong-Xing Yu, Ancong Wu, Wei-Shi Zheng

# Letter-Level Online Writer Identification

Zelin Chen[1] · Hong-Xing Yu[1] · Ancong Wu[1] · Wei-Shi Zheng[1]

**Abstract**
Writer identification (writer-id), an important field in biometrics, aims to identify a writer by their handwriting. Identification in existing writer-id studies requires a complete document or text, limiting the scalability and flexibility of writer-id in realistic applications. To make the application of writer-id more practical (e.g., on mobile devices), we focus on a novel problem, *letter-level online writer-id*, which requires only a few trajectories of written letters as identification cues. Unlike text-\ document-based writer-id which has rich context for identification, there are much fewer clues to recognize an author from only a few single letters. A main challenge is that a person often writes a letter in different styles from time to time. We refer to this problem as the variance of online writing styles (Var-O-Styles). We address the Var-O-Styles in a capture-normalize-aggregate fashion: Firstly, we extract different features of a letter trajectory by a carefully designed multi-branch encoder, in an attempt to capture different online writing styles. Then we convert all these style features to a reference style feature domain by a novel normalization layer. Finally, we aggregate the normalized features by a *hierarchical attention pooling* (HAP), which fuses all the input letters with multiple writing styles into a compact feature vector. In addition, we also contribute a large-scale LEtter-level online wRiter IDentification dataset (LERID) for evaluation. Extensive comparative experiments demonstrate the effectiveness of the proposed framework.

**Keywords** Online writer identification · Online writer identification dataset · Hierarchical Pooling

## 1 Introduction

Writer identification (writer-id) aims to identify a certain writer from a given group of candidates by their handwriting. Writer-id, an important task in the field of security systems, e.g., criminal justice systems and bank account verification systems, is a supplementary approach for biometrics recognition, especially when other devices are not available. The popularity of electronic handwriting devices like smartphones has promoted the development of writer-id and made it important and meaningful in practical scenarios.

However, existing writer-id methods (Namboodiri and Gupta 2006; Gargouri et al. 2013; Dwivedi 2016; Li and Tan 2009; Singh and Sundaram 2015; Venugopal and Sundaram 2017; Li et al. 2007) are restricted to text-\ document-level analysis (i.e., they require a long text or a full document for identification), which limits their scalability and flexibility and thus hinders their practical application. For example, on a smartphone, it is very inconvenient to write a full document.

To further facilitate the applications of writer-id, we propose considering *letter-level online writer-id*, which requires only a few written letters (each of which is an online point trajectory\sequence) for identification. This problem setting is very meaningful in realistic scenarios because it largely promotes the practicability of writer-id. We can conveniently leverage this reliable biometric method for anyone with a touch-screen smartphone by writing several letters. A brief illustration of *letter-level online writer-id* is shown in Fig. 1.

✉ Wei-Shi Zheng
  wszheng@ieee.com

  Zelin Chen
  chenzl9@mail2.sysu.edu.cn

  Hong-Xing Yu
  xKoven@gmail.com

  Ancong Wu
  wuancong@mail2.sysu.edu.cn

[1] School of Data and Computer Science, Sun Yat-set University, Guangzhou, China

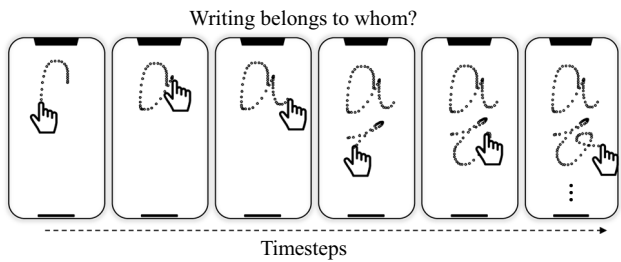

**Fig. 1** Illustration of our *letter-level online writer-id* problem setting, which requires only several trajectories of written letters. The setting is very meaningful in realistic scenarios. For example, we can conveniently leverage writer-id for anyone with a touch-screen smartphone using several written letters and embed the writer-id into a password system

However, *letter-level online writer-id* is challenging due to the large intra-class discrepancy. Specifically, one can handwrite a letter with different speeds, heaviness and shapes, as shown in Fig. 3. We propose to address this variation of online writing style (Var-O-Styles) in a *capture-normalize-aggregate* fashion. Our model aims to *(1)* capture multiple online writing style features from a single input letter trajectory, *(2)* normalize the different style features, and *(3)* aggregate the features to produce a compact one. Specifically, we propose a novel deep network with the following key designs:

(1) We carefully design a multi-branch encoder to capture the characteristics of different online writing styles from a single input letter trajectory.
(2) We propose a letters and styles adapter (LSA) to convert all these style features to a reference feature domain.
(3) We propose a hierarchical attention pooling (HAP) to aggregate rich intermediate features. For each input letter trajectory, HAP fuses writing styles (attentive to a representative style) and the writing temporal information (attentive to distinct segments). Furthermore, HAP merges all input letters according to the reliability of each written letter.

While only closed-set evaluation (testing IDs are seen during training) is considered for the writer-id problem in most existing studies, we additionally conduct an open-set evaluation in which the testing IDs are unseen during training. In the open-set setting, we can directly use our system to identify the unseen writers without retraining. This open-set setting is more practically meaningful. In addition, we also contribute a large dataset for LEtter-level wRiter-ID (LERID) for more evaluation. Our proposed deep framework achieves remarkable performance on LERID, with **99.4**% rank-1 accuracy in the closed-set setting and **93.8**% rank-1 accuracy in the open-set setting.

## 2 Related Work

### 2.1 Writer Identification

Many studies and methods focused on handwriting analysis have been reported during the past three decades, and most of these methods have achieved promising results.

Conventionally, most researches on writer-id are focused on document-level and text-level analysis. Some methods (Tsai and Lan 2005; Namboodiri and Gupta 2006; Bulacu and Schomaker 2007; Schlapbach and Bunke 2007; Li et al. 2007; Schlapbach et al. 2008; El Abed et al. 2009; Tan et al. 2010; Chaabouni et al. 2011; Ramaiah et al. 2013; Shivram et al. 2013; Bertolini et al. 2013; Dwivedi 2016; Venugopal and Sundaram 2018; Lai and Jin 2019; Khan et al. 2018; Venugopal and Sundaram 2018, ?; Nguyen et al. 2019) exploit sophisticated hand-crafted features to model handwriting. For example, *El et al.* (El Abed et al. 2009) modelled writing with temporal sequences and shape codes; (Venugopal and Sundaram 2018) modelled writing based on a codebook extracted by clustering the point and sub-stroke descriptions; (Lai and Jin 2019) modelled writing based on path signature features (Chen 1958); and (Khan et al. 2018) modelled writing with visual descriptors (e.g., SIFT (Arandjelović and Zisserman 2012)). Other techniques (Tang and Wu 2016; Xing and Qiao 2016; Yang et al. 2016; Nasuno and Arai 2017; Zhang et al. 2017; Nguyen et al. 2019; Chen et al. 2018; Christlein and Maier 2018) applied neural networks to model handwriting. For example, (Xing and Qiao 2016) and (Yang et al. 2016) applied a CNN-based encoder to extract handwriting features based on images, and (Zhang et al. 2017) applied RNN to extract features from online sequences. Most deep models apply some hand-crafted data argument method, such as DropSegment (Yang et al. 2016), random permutation (Nguyen et al. 2019) and random hybrid strokes (Zhang et al. 2017). The above studies achieve promising performance at the text- and document-level, but they are not suitable at the letter-level because they do not consider the challenges of *letter-level online writer-id*, such as varying writing styles and different characteristics of letters.

Additionally, the letter-level writer-id is different from the signature verification (Sae-Bae and Memon 2014; Xing et al. 2018; Vorugunti et al. 2019; Lai et al. 2020). In our setting, the content of the writing includes a few predefined letters, whereas in signature verification, the content is without constraint. The uniform content of writing in our setting forces the system to focus on and learn the ID-discriminative writing clues, thus improving the reliability of the system. Moreover, the system can leverage predefined letter characteristics while maintaining practicability on mobile devices.

This work is based on our preliminary work (Chen et al. 2018). The major differences between this work and our preliminary work are as follows:

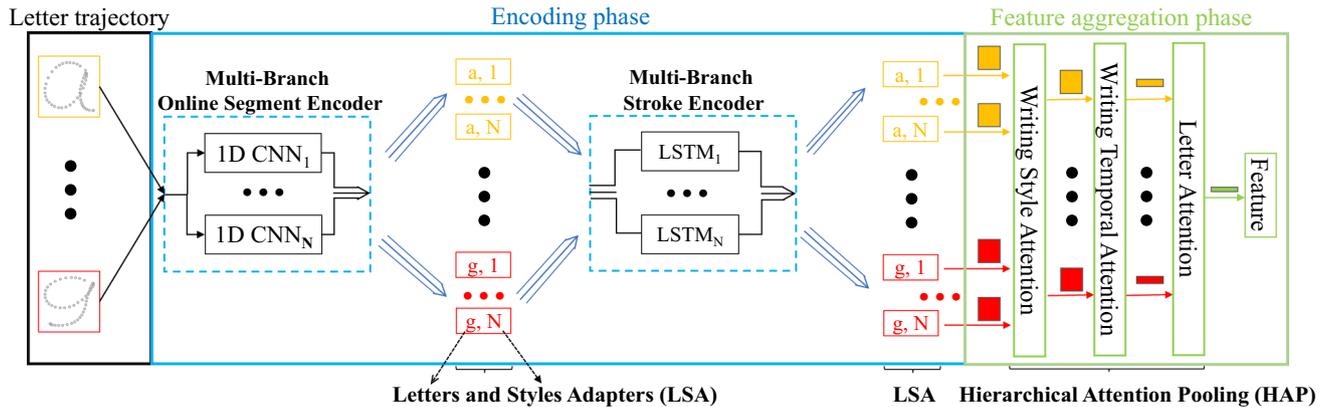

**Fig. 2** The architecture of our proposed deep framework. The input of the framework is several point sequences, each of which corresponds to a letter. Take the yellow letter "a" for example. The "a" goes through a multi-branch encoder which extracts $N$ (the number of branches) features for "a". The LSA layer normalizes the $N$ features of "a" to a reference domain (LSA parameters are not shared across letters). These normalized features are then aggregated by HAP module (writing style attention and writing temporal attention). Finally, for different letters (e.g., "a" to "g"), HAP merges them according to reliability. The details of HAP are shown in Fig. 6

(1) Our preliminary work simply adopted max pooling for feature aggregation, thereby abandoning considerable useful information for identification. To extract useful information from the different features obtained by multiple branches that address different online writing styles, we design a novel HAP method to aggregate the useful information.
(2) Our preliminary work ignored the significance of letter-specific characteristics. In this work, letter-specific characteristics are addressed by LSA to specifically select the significant features of different letters.
(3) We extend the dataset (LetWriterDB) contributed in our previous work and build a LEtter-level online wRiter-ID dataset (LERID), which has approximately 6 times the number of writers of the previous one.

## 2.2 Attention Mechanism

Recently, the effectiveness of attention mechanisms has been demonstrated in many tasks (Nam et al. 2017; Dhingra et al. 2017; Lin et al. 2017; Song et al. 2017; Fu et al. 2019; Si et al. 2018). However, no work has introduced an attention mechanism for writer-id. Generally, an attention mechanism in a neural network can be regarded as an adaptively weighted summation of features, where the weights are dynamically determined by the input. In image classification, the weighted summation is applied along the spatial and channel (in the CNN) dimensions (Wang et al. 2017; Fu et al. 2019; Si et al. 2018). Furthermore, weighted summation can be applied along the temporal dimension in tasks involving a sequence, such as action recognition and natural language processing (Nam et al. 2017; Song et al. 2017; Lin et al. 2017). In contrast, our HAP focuses on different types of ID-discriminative cues that are particularly important for writer-id, including writing styles, writing temporal information and letter aggregation.

## 3 Our Model

### 3.1 An Overview

The goal of our model is to learn ID-discriminative features from a few letters, each of which is an online writing trajectory. The main obstacle is the Var-O-Styles problem, for which we propose the capture-normalize-aggregate fashion. We encourage our model to capture the letter characteristics of different online writing styles, reduce the discrepancies among different styles for each letter type, and finally aggregate them to produce a compact feature.

We show an overview of our model in Fig. 2. In the encoding phase shown in Fig. 2, we design a multi-branch encoder and letters and styles adapter (LSA) to capture the style variation. Specifically, we aim to describe a single letter by $N$ features, each of which is expected to correspond to a representative style. In the feature aggregation phase, we propose hierarchical attention pooling (HAP) to merge these features to obtain a robust style-aware letter feature. HAP also aggregates all input letters into a compact feature discription for writer identification.

### 3.2 Multi-Branch Encoder

Let $\mathbf{X} \in \mathbb{R}^{T \times 2}$ be an online point trajectory\sequence of a single letter, which is the collected raw data (see Fig. 2), where each row $\mathbf{X}(i, :)$ records the spatial coordinates of the $i^{th}$ point and $T = 64$ is the total number of timesteps. We directly use the raw point sequence as the input for the multi-branch encoder instead of a converted image, because it retains the temporal information for extracting online writing style information and is probably more reliable when only a few letters are available. The output of the multi-branch encoder is formulated as

$$[\mathbf{o}_{branch1}, \ldots, \mathbf{o}_{branchN}] = [f_1(\mathbf{X}), \ldots, f_N(\mathbf{X})], \quad (1)$$

where $N$ is the branch number and $f_1, \cdots, f_N$ are encoders with the same network architecture that consume temporal data, e.g., 1D CNN; however, the weights are unshared and differently initialized. By learning different transformations in different branches, we expect to learn to "discretize" online writing styles into $N$ distinct elementary styles, and for each of these styles, we extract a specific feature by a specific branch. Then, we aggregate these $N$ features using an attentional pooling method (described in Sect. 3.4.1). We empirically find that different branches learn differently (see Fig. 12) and characterize different styles (see Fig. 13).

In the following, we elaborate the network architecture of the encoder branch, which consists of an online segment encoder and a stroke encoder. The online segment encoder is a 1D CNN encoder that extracts the segment features in a single stroke, and the stroke encoder is a bidirectional LSTM (biLSTM) (Schuster and Paliwal 1997; Hochreiter and Schmidhuber 1997) encoder used to integrate the segment features into a whole stroke feature. The relationship between the segment encoder and the stroke encoder in a single branch is shown in Figs. 4.

*Online segment encoder* We adopt the 1D CNN (Feng et al. 2015) to extract the segment features in a single stroke. The segment feature is a temporally short-term feature of the current point and its temporally nearby points along the writing trajectory. Specifically, the output of a 1D CNN layer $\mathbf{o} = Conv1D(\mathbf{W}, \mathbf{X})$ (Feng et al. 2015) is defined by

$$o_t = \sum_{i=1}^{S} \sum_{j=1}^{2} w_{i,j} x_{i+t-1,j}, \ t \in \{1, \cdots, T-S+1\}, \quad (2)$$

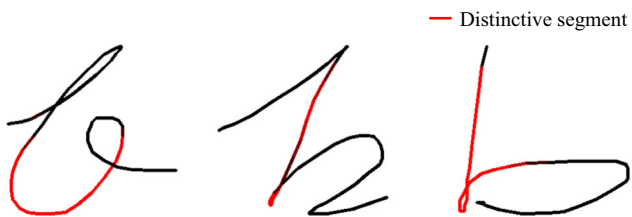

**Fig. 3** An example of the variance of online writing styles (Var-O-Styles) problem is described as follows: the same writer writes the letter "b" in different ways, and the distinctive segments (see Sect. 3.4.2) are different among different online writing styles

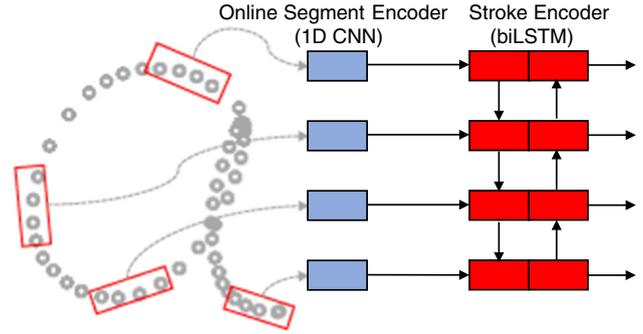

**Fig. 4** Illustration of the relationship between the online segment encoder and the stroke encoder in a single branch. The input of the online segment encoder is the segment sequences (red rectangles in letter "a"), where the length of segments is determined by the kernel size of the 1D CNN. The input of the stroke encoder is the output of the segment encoder, integrating the segment features into a whole stroke feature

where $S = 7$ is the kernel size (i.e., the segment length), $o_t$ is the $t^{th}$ entry of the vector $\mathbf{o} \in \mathbb{R}^{T-S+1}$, and $w_{i,j}$ and $x_{i,j}$ are the entries of the weight matrix $\mathbf{W} \in \mathbb{R}^{S \times 2}$ and the input matrix $\mathbf{X} \in \mathbb{R}^{T \times 2}$, respectively.

*Stroke encoder* To integrate the segment features into a whole stroke feature, we adopt the bidirectional LSTM (biLSTM) (Schuster and Paliwal 1997; Hochreiter and Schmidhuber 1997) encoder. The biLSTM is able to aggregate segment features from both temporal directions. To capture all the segment features for representing the stroke, we exploit all timestep outputs of the biLSTM.

### 3.3 Letters and Styles Adapter

In writer-id, it is necessary to reduce the significant variance among different online writing styles of different letters. To this end, we propose the letters and styles adapter (LSA). The LSA will be used following both the online segment encoder and the stroke encoder in our architecture (see Fig. 2).

We assume that the encodings of different online writing styles of different letters lie in their specific distributions. Specifically, we have $L \times N$ specific distributions, where $L$ is the number of input letters and $N$ is the number of branches. Then, we cast the variance reduction as a distribution normalization problem (see Sect. 3.3.1). In addition, since every feature dimension is not equally discriminative for different letters (see Fig. 5), we further introduce the letter-specific feature selection in LSA (see Sect. 3.3.2).

#### 3.3.1 Distribution Normalization

To alleviate the intra-class discrepancies caused by different online writing styles of different letters, we first assume that each writing style follows a normal distribution (as different writers write the same letter with the same writing style sim-ilarly, e.g., with similar size and rotation); then, the variance is reduced by shifting the different normal distributions to the standard normal distribution Carlucci et al. (2017).

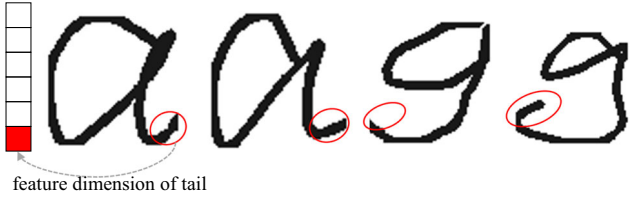

**Fig. 5** The same dimension (written pattern) of the features from a shared letter encoder is not equally discriminative for identification among different letters. For example, the tail in "a" is common so that the feature encoding tail can be less discriminative for identification. In contrast, in writing "g", the tail can be more discriminative as it is less common

Specifically, let **x** be an intermediate encoding of a specific style of a single letter. We denote the covariance matrix and mean vector of **x** as $\sigma^2(\mathbf{x})$ and $\mu(\mathbf{x})$, respectively. We shift all the **x** from different distributions into the same reference distribution by

$$\bar{\mathbf{x}} = diag(\sigma^2(\mathbf{x}))^{-\frac{1}{2}}[\mathbf{x} - \mu(\mathbf{x})], \tag{3}$$

where $diag(\sigma^2(\mathbf{x}))$ is a diagonal matrix with the same diagonal elements as those in $\sigma^2(\mathbf{x})$. In this way, the $N \times L$ distributions are shifted to the same reference distribution.

### 3.3.2 Letter-Specific Feature Selection

Due to the inherent uniqueness and characteristics of letters, the discrimination ability of each feature varies by online writing styles and letters (see Fig. 5). To select the specific significant features for the identification of different letters and styles, specific feature selection strategies are adopted for different letters and styles. Specifically, for each style of a specific letter, we learn a specific $weight$ for feature selection and add a $bias$:

$$\widetilde{\mathbf{x}} = weight \circ \bar{\mathbf{x}} + bias, \tag{4}$$

where "∘" is element-wise multiplication, $weight \in \mathbb{R}^d$ and $bias \in \mathbb{R}^d$ are learnable parameters, and $d$ is the dimension of the feature $\bar{\mathbf{x}}$. In this way, we can select specific significant features in a continuous manner and learn the selection within an end-to-end framework.

Note that, in a single LSA module (see Fig. 3), we have $L \times N$ submodules, each of which corresponds to a specific style (out of $N$ branches) of a letter (out of $L$ input letters). In our implementation, we compute the mean $\mu(\mathbf{x})$ and covariance $\sigma(\mathbf{x})$ in Eq. (3) within each batch during training so that the implementation is similar to the batch-normalization layer Ioffe and Szegedy (2015). However, critically different from batch normalization, $\mu(\mathbf{x})$, $\sigma(\mathbf{x})$ and the parameters $weight$s and $bias$es in LSA are not shared across different styles and letters.

### 3.4 Hierarchical Attention Pooling

We aggregate the rich intermediate features that represent different letters and styles to a compact feature by HAP. Firstly, HAP merges different writing styles of an input letter trajectory. Then it performs a temporal attention pooling to select distinctive segments for temporal aggregation. Finally HAP also fuses all input letters into a compact feature description according to letter reliability. We show an overview of HAP in Fig. 6. The *hierarchical* design (compared to plainly concatenating all features generated by individual attention poolings) reduces computation and leads to a more effectively compact feature.

#### 3.4.1 Writing Style Attention

In the encoding phase, we learn to "discretize" the online writing styles into $N$ distinct elementary styles, and we extract $N$ writing-style-specific features for each letter in a writing-style-agnostic manner. Now, we aggregate these $N$ features for each input letter in a writing-style-aware manner. To this end, the attention generation process should be writing-style-aware (see Fig. 13). At this step, since we focus on the overall writing style of the input letter, we generate the writing style attention by looking at the entire letter. We show an illustration in Fig. 6b.

Specifically, let the outputs of the encoding phase for all written letters be $\{\mathbf{e}^*_{style,i}\}_{i=1}^N$, where $\mathbf{e}^*_{style,i} \in \mathbb{R}^{H \times T}$, $H$ is the feature dimension of the stroke encoder, $T$ is the number of input timesteps, $N$ is the number of branches, and letter denotation $* \in \{a, b, \cdots, g\}$.[1] We extract the image feature $\mathbf{h}^*$ of the corresponding letter image using a 2D CNN image encoder, e.g., VGG (Simonyan and Zisserman 2015). Then, we generate the attention weights by

$$[w^*_{style,1}, \cdots, w^*_{style,N}] = f_{style}(\mathbf{h}^*), \tag{5}$$

where $f_{style}(\cdot)$ is the attention generation function modelled by a fully connected (FC) layer followed by $Softmax(\cdot)$ activation. Then, the writing-style-aware feature aggregation is given by

$$\mathbf{e}^*_{time} = \sum_{i=1}^N w^*_{style,i} \mathbf{e}^*_{style,i}, \tag{6}$$

---
[1] For convenience of description and simplicity of denotation, we assume that the writer writes the specific letters 'a', 'b', ..., 'g'.

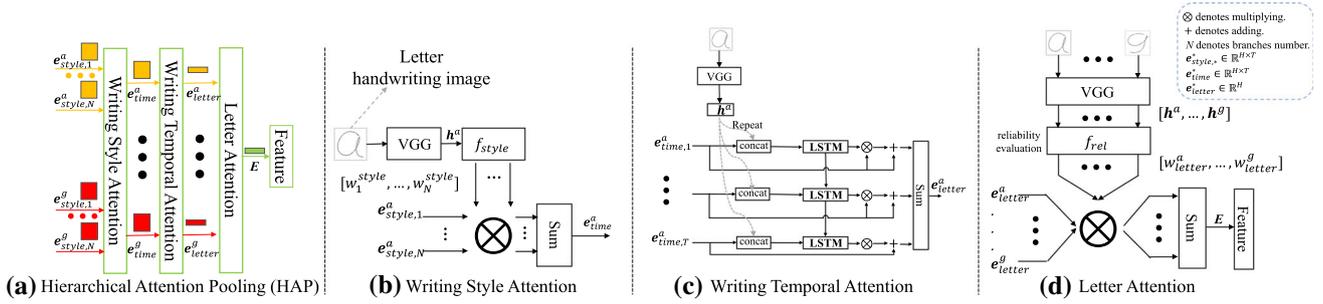

**(a)** Hierarchical Attention Pooling (HAP)  **(b)** Writing Style Attention  **(c)** Writing Temporal Attention  **(d)** Letter Attention

**Fig. 6** The details of the hierarchical attention pooling (HAP). **b**, **c** is an example of "Writing Style Attention"/"Writing Temporal Attention" of letter 'a'. Please refer to the text in Sect. 3.4 for details

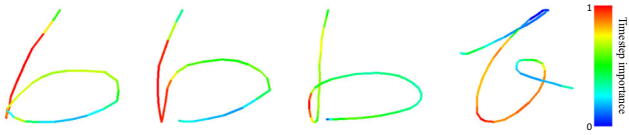

**Fig. 7** The writing temporal attention. For visualization, the attention is normalized across the timesteps. For specific writing, the timesteps in red are more important than the timesteps in blue. The results show that the model can find the most distinct timesteps (segments) for identification, as the parts highlighted in red are the most important timesteps (segments) for distinguishing the four writers (Color figure online)

where we call $e^*_{time} \in \mathbb{R}^{H \times T}$ the temporal feature, which will be the input of the next temporal attention hierarchy.

### 3.4.2 Writing Temporal Attention

For specific online writing, not all timesteps for writing a letter are equally distinct for writer-id (see Fig. 7). For example, the most important timesteps should appear in the online segment that can distinctly identify a specific writer. The main purpose of the writing temporal attention is to emphasize the most important timesteps for constructing letter features. To this end, we introduce an distinctive-segment-aware attention generation process, where the corresponding holistic image of the handwriting is leveraged to guide the generation of timestep importance under a global written context. We show an illustration in Fig. 6c.

Specifically, we model the attention generation function by using an LSTM that can automatically learn the relationship within the trajectory. The attention generation input $g^*_t$ for each timestep $t$ is the concatenation of the holistic image feature $h^*$ and the temporal feature $e^*_{time,t}$ at the timestep $t$. In other words, $g^*_t = [h^*, e^*_{time,t}]$, $\forall t = 1, \cdots, T$. Then, the attention weights are given by

$$[w^*_{time,1}, \cdots, w^*_{time,T}] = Softmax(LSTM(g^*_1, \cdots, g^*_T)), \quad (7)$$

where $LSTM(g^*_1, \cdots, g^*_T)$ is a sequence of output scalars generated by the LSTM. The temporal features $e^*_{time}$ are finally aggregated to a letter feature $e^*_{letter}$:

$$e^*_{letter} = \sum_{t=1}^{T} w^*_{time,t} e^*_{time,t}, \quad (8)$$

where $e^*_{letter} \in \mathbb{R}^H$ is a single-letter feature (e.g., "a") after fusing the writing styles and writing temporal information, which will be fed into the next letter attention hierarchy.

In the temporal hierarchy, the important timestep for each written letter is paid more attention by increasing the corresponding weight. Meanwhile, the feature dimension is reduced drastically, (from $HT$ in $e^*_{time}$ to $H$ in $e^*_{letter}$) to produce a more compact feature,[2] which alleviates overfitting and reduces the computational cost when testing.

### 3.4.3 Letter Attention

In practice, a person could write more than one letter online, and a person sometimes writes a letter casually, and therefore, the written letter can be too irregular to contribute to identification (or even harmful and disturbing for a writer-id system). To alleviate the influence of unreliable letters when aggregating all input letters, we introduce letter attention pooling based on letter reliability to improve the robustness of the model. We show an illustration in Fig. 6d.

Specifically, we model the letter reliability evaluation function $f_{rel}(\cdot)$ by using an FC layer that takes the holistic letter image feature $h^*$ ($* \in \{a, b, \cdots, g\}$) as input and outputs a reliability scalar value. Then, the attention weights are given by

$$[w^a_{letter}, \cdots, w^g_{letter}] = Softmax([f_{rel}(h^a), \cdots, f_{rel}(h^g)]). \quad (9)$$

---
[2] In our setting, $T = 64$ and $d = 512$. If we directly flatten $e^*_{time}$, the dimension is over 30k.

The $[w_{letter}^{a}, \cdots, w_{letter}^{g}]$ denotes the letter importance values for identification from the written letters. Finally, all letter features are aggregated by

$$\mathbf{E} = \sum_{*=\{`a',\cdots,`g'\}} w_{letter}^{*} \mathbf{e}_{letter}^{*}, \tag{10}$$

where $\mathbf{E}$ is the final feature for the writer-id task.

## 4 Experiments

In this section, we compare the proposed framework with other related writer-id models and perform extensive ablation studies to further verify the effectiveness of the proposed submodules and the robustness of the full model.

### 4.1 Dataset

We conducted extensive experiments on three online writer-id datasets, namely, Modified IAM (IAM), LetWriterDB (Chen et al. 2018) and a new collected dataset called LERID. *Modified IAM* We modified IAM dataset to make it suitable for the letter-level online writer-id. The primitive IAM dataset contains approximately 200 writers, and the writers wrote different text. The writing examples were recorded online. The modification of the primitive IAM dataset is as follows. We first truncated the letters from the writing text according to the start timesteps and the end timesteps of the letters. To ensure sufficient examples for training and testing, we chose letters and writers such that *(1)* each chosen writer provides at least $m = 60$ examples of each chosen letter type, and *(2)* each chosen letter type is provided by at least $n = 60$ chosen writers. Formally,

$$\mathscr{V}^{\text{wri}}(m, n) = \bigcap_{l \in \mathscr{V}^{\text{let}}(m,n)} \mathscr{B}(l, m), \tag{11}$$

$$\mathscr{V}^{\text{let}}(m, n) = \left\{ l \mid |\mathscr{B}(l, m)| > n, \, l \in \mathscr{A} \right\}, \tag{12}$$

$$\mathscr{B}(l, m) = \left\{ p \mid N(p, l) \geq m, \, p \in \mathscr{P} \right\}, \tag{13}$$

where $\mathscr{V}^{\text{wri}}$ denotes the set of chosen writers, $\mathscr{V}^{\text{let}}$ denotes the set of chosen letter types (the predefined letter types), $\mathscr{A}$ is the alphabet, $\mathscr{B}(l, m)$ denotes the set of writers provided at least $m$ examples of letter '$l$', $N(p, l)$ is the number of the examples of letter '$l$' provided by the writer '$p$' and $\mathscr{P}$ is the set of all writers in IAM. Finally, letter {'a', 'c', 'd', 'e', 'h', 'i', 'l', 'n', 'o', 'r', 's', 't', 'u', 'g'} and a total of 84 writers were chosen.

Some examples of Modified IAM (IAM) are shown in Fig. 8. We proposed two evaluation protocols for Modified IAM (IAM) dataset, *i.e.,* a closed-set protocol and an open-set protocol. The details are as described follows:

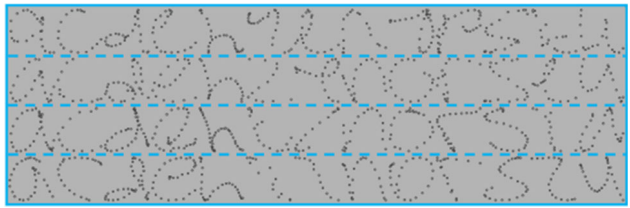

**Fig. 8** Examples of modified IAM (IAM). The selected letters are {'a', 'c', 'd', 'e', 'h', 'i', 'l', 'n', 'o', 'r', 's', 't', 'u'}. Each sample instance is the online point trajectory of a single written letter

*(1)* Closed-set: All the testing IDs are seen during training. The examples of each letter type of each writer were split into a training set and test set with a ratio of 3:1. The statistics are shown in Table 1.
*(2)* Open-set: All the testing IDs are unseen during training. We randomly chose the examples of 63 writers for training and the left examples of 21 writers for testing. There is no overlap of writer IDs between the training set and test set.

*LetWriterDB* The dataset contains 60 writers and 6 letters ({'a', 'b', 'c', 'd', 'e', 'g'}) with over 30k online letters (approximately 21k for training and 9k for testing). The writing examples were collected on mobile phones (with no restriction on brands) from a website (see Fig. 1). We follow the evaluation protocol (closed-set protocol) in Chen et al. (2018), where the examples of each letter type of each writer were split into the training set and test set with a ratio of 2:1. The statistics of LetWriterDB (under the closed-set setting) are shown in Table 1.

*LERID* The LEtter-level online wRiter-ID (LERID) dataset is extended from LetWriterDB (Chen et al. 2018) and has approximately 6 times the number of writers as LetWriterDB. The examples were collected on mobile phones (arbitrary OS and brands) from a website (see Fig. 1). Some collected examples are shown in Fig. 9. The main purpose of the newly collected dataset is to provide more training samples and writer IDs to help the model learn more robust features. During the collection, each writer was asked to provide more than 30 examples of each letter type. Finally, 107,723 online letters and 414 writer IDs were finally collected. We also proposed a closed-set protocol and an open-set protocol for the LERID dataset. The details are as follows:

(1) Closed-set: The examples of each letter type of each writer were split into training set and test set with a ratio of 3:1. The statistics are shown in Table 1.
(2) Open-set: We randomly selected the examples of 314 writers for training and the examples of the rest 100 writers for testing. There is no overlap of writer IDs between the training set and test set.

**Table 1** The statistics of Modified IAM (IAM), LetWriterDB (Chen et al. 2018) and the newly collected dataset LERID for *letter-level online writer-id* under a closed-set setting

| Letter | a | c | d | e | h | i | l |
|---|---|---|---|---|---|---|---|
| Modified IAM (IAM) | | | | | | | |
| Train | 12,729 | 4732 | 6644 | 20,847 | 9352 | 10,857 | 6551 |
| Test | 4273 | 1610 | 2242 | 6983 | 3146 | 3649 | 2215 |
| Letter | n | o | r | s | t | u | Total |
| Train | 10,791 | 12,392 | 9,928 | 10,324 | 14,436 | 4,913 | 1,344,496 |
| Test | 3,623 | 4,158 | 3,340 | 3,472 | 4,838 | 1,671 | 45,220 |
| Letter | a | b | c | d | e | g | Total |
| LetWriterDB (Chen et al. 2018) | | | | | | | |
| Train | 3,730 | 3,474 | 3,339 | 3,242 | 3,254 | 3,237 | 20,276 |
| Test | 1,489 | 1,352 | 1,381 | 1,324 | 1,334 | 1,391 | 8,271 |
| Letter | a | b | c | d | e | g | Total |
| LERID | | | | | | | |
| Train | 14,138 | 13,104 | 13,101 | 13,020 | 12,914 | 13,018 | 79,295 |
| Test | 5,024 | 4,726 | 4,832 | 4,570 | 4,665 | 4,611 | 28,428 |

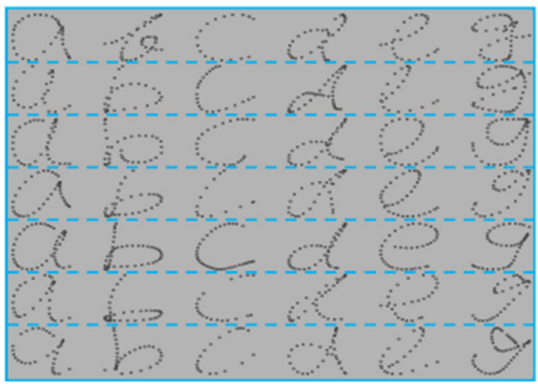

**Fig. 9** Examples of our collected online LEtter-level wRiter-ID (LERID) dataset. Each sample instance is the online point trajectory of a single written letter

### 4.2 Implementation Details

The online segment encoder is a one-layer 1D CNN with $tanh(\cdot)$ activation, and the stroke encoder is a one-layer biLSTM. For VGG, we adopt the vgg5-bn (16-32-32-64-64) architecture. For training, we add an FC layer after HAP (the final feature **E**) and train the deep framework (including the parameters of LSA and HAP) in an end-to-end fashion with norm softmax loss (Wang et al. 2017) to learn a hypersphere feature embedding (Liu et al. 2017). The RMSprop (Tieleman and Hinton 2012) optimizer with a 1e-3 learning rate is adopted, and the learning rate is decayed every 100 epoch (totally 500 epoch) with a decay ratio of 0.5. We do not adopt any data argument method when training. To make the model insensitive to the various screen sizes of different devices and different regions of the screen when the letters are written (i.e., translation invariance), the input coordinate of each letter is normalized to $[-1, 1]$. Unless otherwise stated, we set the branch number $N = 3$, and the evaluation of $N$ is shown in Sect. 4.4.1 (see Fig. 11). We multiply a learnable scalar after the output of LSTM in Eq. (7) to alleviate the effect of *Softmax Bound* (Wang et al. 2017) and add $1/T$ to the $\mathbf{w}^*_{time,i}$ in Eq. (8) to gain a smooth attention.

For the closed-set setting, we directly use the last FC layer for recognition. For the open-set setting, we remove the last FC layer and use the final feature **E** to represent the writing. We randomly selected one sample for each test writer to form the template (namely gallery) set and compute pairwise cosine similarity between the testing example and template examples to evaluate the rank-1 and rank-5 accuracies (Moon and Phillips 2001). We repeated the evaluation 10 times with different template sets and report the mean performance.

Our model (implemented in Pytorch v1.4) can identify a writer by 6 letters in a short time of approximately 20ms when using a TITAN X GPU and 110ms if only using E5-2686 v4 CPU, which is able to meet the requirement of real-time writer-ID systems.

### 4.3 Comparison with State-of-the-Art Methods

For the closed-set setting, we compare our model with two hand-crafted models (point-based (Gargouri et al. 2013) and histogram-based (Dwivedi 2016)) and two deep models (DeepWriterID (Yang et al. 2016) and DeepRNN (Zhang et al. 2017)) on three online writer-id datasets (Modified IAM (IAM), LetWriterDB (Chen et al. 2018) and LERID). The

**Table 2** Comparison with other models under the closed-set setting at rank-1 accuracy

| Methods | Modified IAM (IAM) (%) | LetWriterDB Chen et al. (2018) (%) | LERID (%) |
|---|---|---|---|
| Point-based (Gargouri et al. 2013) (Hand-crafted) | 32.0 | 80.0 | 73.8 |
| Histogram-based (Dwivedi 2016) (Hand-crafted) | 44.0 | 88.8 | 85.0 |
| DeepWriterID (Yang et al. 2016) (2D CNN) | 50.7 | 64.8 | 86.9 |
| DeepRNN (Zhang et al. 2017) (LSTM) | 92.2 | 92.5 | 88.7 |
| Ours | **99.9** | **99.8**% | **99.4** |

The bold values denote that the method achieve the best performance

**Table 3** Comparison with other models under the open-set setting

| Methods | Modified IAM (IAM) | | LERID | |
|---|---|---|---|---|
| | Rank-1 (%) | Rank-5 (%) | Rank-1 (%) | Rank-5 (%) |
| Point based (Gargouri et al. 2013) (Hand-crafted) | 9.7 | 30.0 | 4.2 | 15.1 |
| Histogram based (Dwivedi 2016) (Hand-crafted) | 7.4 | 28.8 | 20.2 | 36.2 |
| DeepWriterID (Yang et al. 2016) (2D CNN) | 52.5 | 85.9 | 68.5 | 87.2 |
| DeepRNN (Zhang et al. 2017) (LSTM) | 57.4 | 93.4 | 58.7 | 82.7 |
| Ours | **76.2** | **96.2** | **93.8** | **99.3** |

The bold values denote that the method achieve the best performance

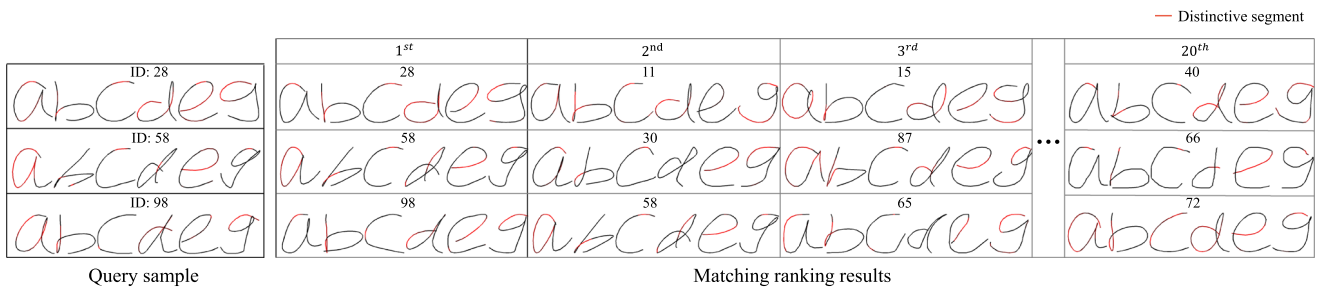

**Fig. 10** Some ranking results of our model for the open-set setting, where the distinctive segments of the writing are highlighted in red (produced by the writing temporal attention in HAP). Although the written letters at the front of the ranking list are very similar, we can find the differences from the distinctive online segments along with the writing trajectory. For example, the writing examples of IDs "28" and "11" in the ranking list at the first row are very similar, but they can be distinguished from the distinctive online segments of the letter "g" (Color figure online)

feature of DeepWriterID (Yang et al. 2016) is the 2D CNN feature extracted from the letter path-signature (Chen 1958) feature maps. The feature of DeepRNN (Zhang et al. 2017) is the bidirectional LSTM feature extracted from writing trajectories. The rank-1 accuracy comparison of the models is shown in Table 2.

For the open-set setting, we compare our model with these models on Modified IAM (IAM) and LERID. The comparison of the rank-1 and rank-5 accuracies of the models is shown in Table 3. We also show some ranking results on LERID in Fig. 10.

*Comparison with hand-crafted models* Our model outperforms the conventional hand-crafted feature-based models by a large margin, e.g., the improvement is greater than 73%\ 68% at rank-1 compared to the *Histogram-based method* (Dwivedi 2016) under the open-set setting on LERID\ Modified IAM (IAM). The results imply that the *letter-level online writer-id* is challenging. Our model can extract the writing clue embedding at the letter-level much better than the hand-crafted models.

*Comparison with deep models* Compared with the deep models, the improvement of our model is clear, e.g., the improvement is over 25%\ 23% at rank-1 compared with *DeepWriterID* (Yang et al. 2016) under the open-set setting on LERID\ Modified IAM (IAM). The main reasons for the improvement are as follows. In LSA, the model takes Var-O-Styles and letter-specific features into consideration. In HAP, the model leverages attention pooling to maintain the important information while weakening unimportant and harmful information and aggregating information hierarchically. It thus makes use of the information in a sufficient and effective manner. The compared methods do not have the above characteristics and are therefore inferior.

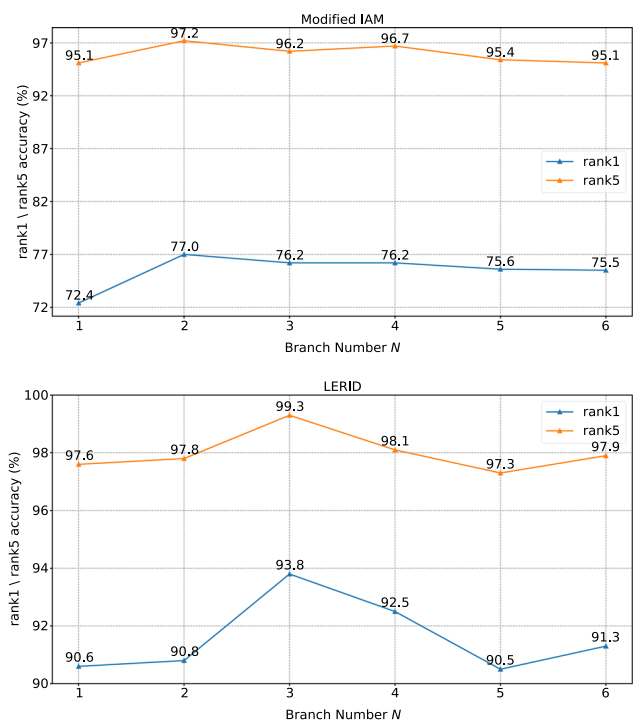

**Fig. 11** The effect of different numbers of branches on Modified IAM (IAM) and LERID

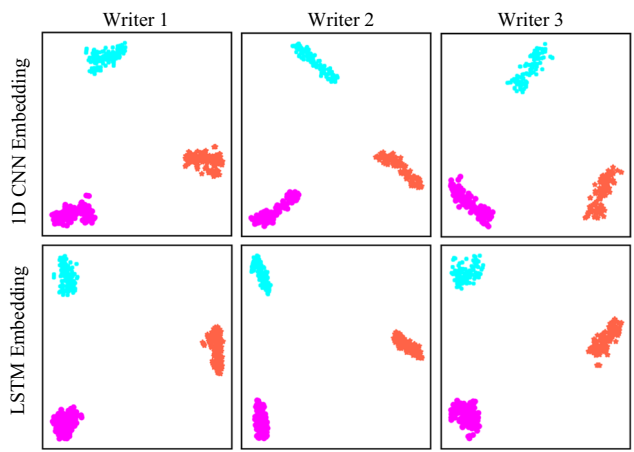

**Fig. 12** T-SNE Maaten, L.v.d., Hinton, G. (2008) visualization of the embedding features of the letter "a" written by the same writer on LERID. Points with the same colour represent the output of the same branch, and each column denotes a specific writer. The result shows that, even without the constraints, different branches can learn differently.

### 4.4 Ablation Study

In the following, the ablation study of our method is performed on Modified IAM dataset (IAM) which has a large variety of letter's types and the large-scale LERID dataset.

#### 4.4.1 Different Numbers of Branches

To evaluate the impact of using multiple branches, we compare the performance of our model with various values of $N$. As shown in Fig. 11, the multi-branch encoder handles the multi-modal inputs better (with a 4.0% improvement at rank-1 on LERID compared with using only one branch (i.e., $N = 1$)). Specifically, $N = 3$ branches can give empirically satisfactory performances. This might be because the handwritten letter trajectories follow a tri-modal-like distribution which can be better modeled by 3 branches than a single branch. The significant improvement shows that using the multi-branch design is effective.

We visualize the feature embeddings on LERID of the segment encoder (1D CNN) and the stroke encoder (LSTM) in different branches. As shown in Fig. 12, the embedding features of different branches form different clusters.

This qualitative result implies that the different branches could learn to extract different features for a single input letter trajectory. Although it is difficult to quantitatively measure the writing styles, we can expect the branches to capture different styles. In Sec. 4.4.3 we can also see the connection between the features and styles in Fig. 13.

#### 4.4.2 Ablation Study on LSA

We show ablation study results in Table 4. Without letter-specific feature selection (see Eq. (4)), the performance drops on both IAM and LERID datasets. This observation implies that we should consider the inherent uniqueness and characteristics of different letters when encoding. If we further remove the distribution normalization operation (see Eq. (3)), the performance drops significantly on both datasets. This result shows that shifting the different distributions to the standard normal distribution is an essential modeling component, probably because it could alleviate the Var-O-Styles problem. Overall, comparing "Full model" to "w/o LSA", the remarkable performance gap demonstrates the effectiveness of the proposed LSA layer.

*Comparison to different letter adapter strategies* To show the benefits of adopting specific adapters for different styles and letters, we explored different partially-specific (partially-sharing) strategies of the letter adapters as shown in Table 5. The strategies of specific adapters, including the style-specific ("Letter-sharing"), letter-specific ("Style-sharing") and style-letter-specific (LSA), outperform the strategy of universal adapter ("All-sharing") noticeably, e.g., an improvement of 44.3% at rank-1 is achieved by replacing universal adapter ("All-sharing") with the style-letter-

**Table 4** Ablation study of LSA

| Methods | Normalization | Selection | Modified IAM (IAM) | | LERID | |
|---|---|---|---|---|---|---|
| | | | Rank-1 (%) | Rank-5 (%) | Rank-1 (%) | Rank-5 (%) |
| w/o LSA | × | × | 64.5 | 91.4 | 85.4 | 95.5 |
| w/o selection | ✓ | × | 75.2 | 95.0 | 92.1 | 98.1 |
| Full model | ✓ | ✓ | **76.2** | **96.2** | **93.8** | **99.3** |

The bold values denote that the method achieve the best performance

**Table 5** Comparison to different partially-specific (partially-sharing) strategies of the adapters

| Methods | Style-specific | Letter-specific | Modified IAM | | LERID | |
|---|---|---|---|---|---|---|
| | | | rank-1 (%) | rank-5 (%) | rank-1 (%) | rank-5 (%) |
| All-sharing | × | × | 31.9 | 65.8 | 75.6 | 90.2 |
| Letter-sharing | ✓ | × | 71.6 | 94.6 | 91.0 | 98.3 |
| Style-sharing | × | ✓ | 75.4 | 94.9 | 92.3 | 98.3 |
| Ours (LSA) | ✓ | ✓ | **76.2** | **96.2** | **93.8** | **99.3** |

The bold values denote that the method achieve the best performance

**Table 6** Ablation study on HAP

| Methods | Attention type | | | Modified IAM (IAM) | | LERID | |
|---|---|---|---|---|---|---|---|
| | Style | Temporal | Letter | rank-1 (%) | rank-5 (%) | rank-1 (%) | rank5 (%) |
| Mean pooling | × | × | × | 72.4 | 91.8 | 85.5 | 96.0 |
| Max pooling | × | × | × | 67.8 | 92.7 | 90.3 | 97.3 |
| w/ style | ✓ | × | × | 73.8 | 94.3 | 91.2 | 98.0 |
| w/ style & temporal | ✓ | ✓ | × | 75.5 | 95.3 | 92.7 | 98.7 |
| Order changed | ✓ | ✓ | ✓ | 75.8 | 96.1 | 93.6 | 98.7 |
| Full model | ✓ | ✓ | ✓ | **76.2** | **96.2** | **93.8** | **99.3** |

The "Mean\ Max pooling" means HAP is replaced with mean\ max pooling. For "w/ style", the temporal attention and letter attention are replaced with mean pooling. For "w/ style & temporal", the letter attention is replaced with mean pooling. The "Order changed" means the hierarchical positions of the style attention and the temporal attention are exchanged.
The bold values denote that the method achieve the best performance specific (LSA)); and the style-letter-specific (LSA) strategy also outperforms the partially-specific strategies (*i.e.,* the style-specific and letter-specific). Such result implies that designing specific adapters for different styles and letters is advisable due to the differences (gaps) of distribution and discriminative features of different styles and letters.

### 4.4.3 Ablation Study on HAP

We perform the ablation study on HAP in Table 6.
*Effect of HAP* Comparison of the "Mean\ Max pooling" and the "Full model" shows the effectiveness of HAP (a 3.8%\ 8.4% improvement at rank-1 on Modified IAM (IAM) and an 8.3%\ 3.5% improvement at rank-1 on LERID), which indicates that hierarchical attention pooling can aggregate rich intermediate feature encodings more effectively than the simple pooling strategies and thus extract better compact features.

*Effect of writing style attention* By adopting the writing style attention, the performance ("w/ style" in Table 6) is much better than that of the "Mean pooling", which implies that the specific branch should be selected via writing style attention. The attention on different writing styles is illustrated in Fig. 13. Figure 13 shows that the same letter type "a" can receive different attention weight distributions according to the hand-written style. This indicates that the writing style attention module is indeed style-aware and it merges the features in a reasonable way.

*Effect of temporal attention* By additionally adopting the writing temporal attention, the performance of the "w/ style & temporal" is better than that of the "w/ style" only, which implies that temporal attention can alleviate the effect that mean pooling among timesteps weakens the discriminative timesteps. This is achieved by assigning smaller attention weights to the timesteps that carry little discriminative information. We also show some examples of the writing temporal attention on LERID in Fig. 7.

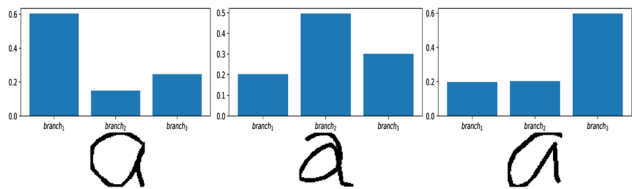

**Fig. 13** Attention weights on different branches on LERID. The y-axis denotes the attention outputs, and the x-axis denotes the branches representing different writing styles

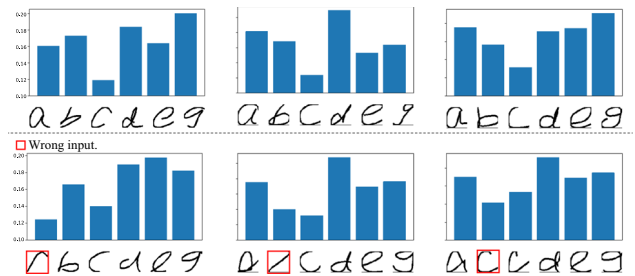

**Fig. 14** The letter attention of some correctly identified samples in the testing set on LERID. The y-axis denotes the letter attention values, and the x-axis denotes different letters. The first row is the clean data, and the second row is the dirty data with some wrong letter inputs. Our model can robustly identify the writer, even with a few wrong\ unclear letters, since the model ignores the wrong letter by assigning less attention to it.

*Effect of letter attention* The performance of our "Full model" is better than that of the "w/ style & temporal", which implies that the network with the extra letter attention is more robust because the writer is identified based on more reliable letters.

As mentioned in Sect. 3.4.3, the letter attention weights in Eq. (9) imply the importance of each letter. To verify that the network can automatically evaluate letter importance, even without explicit supervision, we show some correctly identified samples with their letter attentions in Fig. 14. The first row shows the clean data, and the second row shows some dirty data with wrong letter input. As shown in the first row in Fig. 14, the model clearly gives low importance to the letter "c" compared to the other letters since the letter "c" has less distinct writing characteristics than the other letters. Furthermore, the letter attention makes the model more robust, as shown in the second row in Fig. 14, where the wrong letter input is given lower importance compared to the usual case (i.e. the first row). The result shows that, even with wrong\ unclear input in the sample, our model can still identify the writer by other reliable letters.

*The order of hierarchies* The writing style hierarchy is placed before the writing temporal hierarchy in HAP for computational efficiency (reduced by at least 1.2 GFLOPs for each identification), since the computational complexity in the temporal hierarchy is much greater than that in the writing style hierarchy.

The performance after exchanging the order of the writing style hierarchy and the temporal hierarchy is shown in the "Order changed" in Table 6, which is slightly worse than that of the "Full model" (0.2%\ 0.4% on LERID\ Modified IAM (IAM) at rank-1). The increased number of parameters (as each style needs a specific temporal attention in "Order changed") and decreased performances support our HAP design. The adopted order in our method is more effective and efficient.

#### 4.4.4 Effect of the Combination of LSA and HAP

We further study the combination of LSA and HAP and show the results in Table 7. The performance comparison of the "w/o LSA, HAP" and the "w/o HAP" (Max pooling) shows that the model gains little improvement by solely applying LSA. The main reason is that, max pooling can alleviate the problem of Var-O-Style to some limited extent without HAP (Chen et al. 2018). Also, solely applying HAP would hamper the model's performance (compared the "w/o LSA, HAP" with the "w/o LSA"), because without LSA, HAP aggregates "absolute" features from different distributions and ignores the relative relationships. Therefore, the combination of LSA and HAP is necessary in our framework.

#### 4.4.5 Robustness of the Proposed Model

*Writing fewer letters* The motivation of our work is to make writer-id easier to use on mobile devices. In our proposed model, we can further reduce the number of letters while maintaining the performance. We traverse all the combinations of two, three, four, five and six letters on LERID. For example, when adopting two letters, the combinations are "ab", "ac", ..., "eg", for a total of 15 combinations. Then, the model is retrained in order to learn a new letter attention pooling layer for each instance. The mean performance when adopting two, three, four, five and six letters is reported in Table 8. With fewer letters, our model can still identify writers with remarkable performance, which demonstrates the high practicability of our model.

*Permuting letters* We randomly permute the letter order of the test samples, rank the order again by a trained letter classifier (99.8% accuracy on the letter classification task), and then feed the re-ranked samples into our model. We achieve a rank-1 performance of 93.3% under the open-set setting. This performance is comparable to that of the model without permutation (rank-1 93.8%). Therefore, our model can be embedded into a password system. Such a system could

**Table 7** The effect of the combination of LSA and HAP

| Methods | LSA | HAP | Modified IAM (IAM) | | LERID | |
|---|---|---|---|---|---|---|
| | | | rank-1 (%) | rank-5 (%) | rank-1 (%) | rank-5 (%) |
| w/o LSA, HAP | × | × | 68.5 | 93.3 | 90.8 | 97.7 |
| w/o HAP | ✓ | × | 67.8 | 92.7 | 90.3 | 97.3 |
| w/o LSA | × | ✓ | 64.5 | 91.4 | 85.4 | 95.5 |
| Full model | ✓ | ✓ | **76.2** | **96.2** | **93.8** | **99.3** |

"w/o LSA, HAP" means that LSA is not adopted and HAP is replaced by max pooling (Chen et al. 2018).
"w/o HAP" replaces HAP with max pooling
The bold values denote that the method achieve the best performance

**Table 8** The performance of writing fewer letters on LERID

| #Letters | 6 (%) | 5 (%) | 4 (%) | 3 (%) | 2 (%) |
|---|---|---|---|---|---|
| rank-1 (closed-set) | 99.4 | 99.5 | 99.0 | 97.0 | 90.2 |
| rank-1 (open-set) | 93.8 | 90.3 | 86.1 | 79.9 | 58.0 |
| rank-5 (open-set) | 99.3 | 97.6 | 96.0 | 93.0 | 78.2 |

"#Letter" denotes the number of adopted letters

**Table 9** The performance under the letter-independent setting

| #Letter | 2 (%) | 3 (%) | 4 (%) | 5 (%) | 6 (%) |
|---|---|---|---|---|---|
| Modified IAM (IAM) (closed-set) | | | | | |
| rank-1 | 89.4% | 96.0 | 98.6 | 99.4 | 99.8 |
| LERID (closed-set) | | | | | |
| rank-1 | 94.0 | 98.4 | 99.4 | 99.7 | 99.8 |
| LERID (open-set) | | | | | |
| rank-1 | 51.7 | 62.6 | 75.3 | 82.6 | 93.8 |
| rank-5 | 72.7 | 81.1 | 89.3 | 93.7 | 98.4 |

"#Letter" denotes the number of adopted letters be reliable since a hacker would need to crack the password and imitate personal handwriting at the same time.

*Evaluation on a letter-independent setting* Our framework can also be applied to letter-independent setting (as letter types are limited), where the writer can write any letter's types (but limited to the letter's types registered at training) during identification. To this end, for training (both in the closed-set and the open-set), we randomly drop some letters' features ($e^*_{letter}$) and their corresponding letter attention logit ($f_{rel}(h^*)$) before $Softmax(\cdot)$. Next, we recompute the aggregated feature **E** with the remaining letters using Eqs. (9) and (10) and adopt such features to compute the loss. Such a strategy can be considered a regularization to make the model more robust, and the model can still be trained in an end-to-end fashion.

The performances under the letter-independent setting are shown in Table 9. For the closed-set setting, we randomly chose 2\ 3\ 4\ 5\ 6 letters to identify the writers. For the open-set setting, we traverse all combinations of two, three, four, five and six letters (on LERID) during testing as described in "Writing fewer letters" and report the mean performance; however, we train only one model (without retraining) this time. The performance of our model can still achieve high accuracy without much decay compared with the letter-dependent setting, which implies that our model can be easily applicable to more the realistic setting, i.e., the letter-independent setting.

*Robustness to different predefined letters* To explore the influence of choosing different predefined letters, we evaluated our model on different (disjoint) letter groups. The experiment details are as follows. To choose more predefined letters, we first set $m = 30$ (the least number of examples of a letter of a writer ID) and $n = 60$ (the least number of shared writers of a letter type) as in Eq. (12), (13); then 20 letters ({'a', 'b', 'c', 'd', 'e', 'f', 'g', 'h', 'i', 'l', 'm', 'n', 'o', 'p', 'r', 's', 't', 'u', 'w', 'y'}) and a total of 88 writer IDs were chosen. Next, we divided the chosen letter types into 4 disjoint groups ({'a', 'b', 'c', 'd', 'e'}, {'f', 'g', 'h', 'i', 'l'}, {'m', 'n', 'o', 'p', 'r'} and {'s', 't', 'u', 'w', 'y'} respectively) and validated our model on the 4 groups. Each group datasets was formed by sampling $m$ examples of each chosen letter type and ID to ensure that the 4 datasets consisted of the same amount of data of each letter and ID. For open-set setting, we randomly chose the examples of 66 writers for training and left 22 writers for testing. For closed-set setting, the examples of each letter type of each writer were randomly split into training set and testing set with a ratio of 3:1. The performances of different letter groups are shown in Table 10, where the performance variances are relatively small (*e.g.,* less than 1% at rank-1 (closed-set)). Such result implies that our model is not sensitive to the predefined letter types.

*Reject the hackers input unseen letters* To explore the ability of our model to reject the hackers that input unseen letters, we feed different unseen letters (with respect to the training letters) into the model and the evaluation performance at rank-1 (closed-set) are shown in Fig. 15. The adopted datasets of the experiment are the same as those described in "Performance of different predefined letters". Clearly, the performance decreased a lot (at least 40%) if unseen letters are fed into the model. The main reason of the reduction is that the letter specific adapters (LSA) are specific for each registered letter and is not designed to adapt the unseen letters. Such property further ensures the safety of the writing password system that verifies the password (by a letter classification system) and writing (by a writer-ID system) at the same time. Specifically, considering that someone (the hackers) input some unseen letters into the system, the writer-ID system (adopting our model) is still able to reject the IDs (as the ID recognized by the model differs from the login ID) even when the letter classification system failed (have been hacked).

**Table 10** Results of different predefined letters, which shows that our model is not sensitive to the predefined letters

| Letters | a,b,c,d,e (%) | f,g,h,i,l (%) | m,n,o,p,r (%) | s,t,u,w,y (%) |
|---|---|---|---|---|
| rank-1 (open-set) | 38.1 | 37.0 | 37.1 | 37.4 |
| rank-5 (open-set) | 66.4 | 67.2 | 70.5 | 67.4 |
| rank-1 (closed-set) | 85.7 | 85.4 | 85.4 | 86.2 |

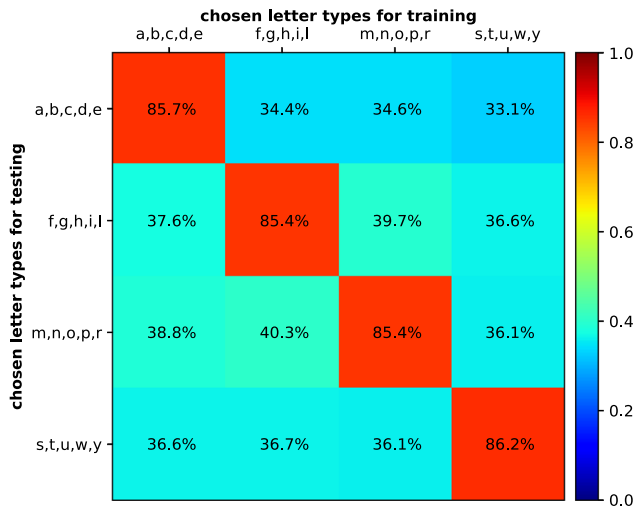

**Fig. 15** Results (at rank-1 (closed-set)) of different chosen letter types for training and testing

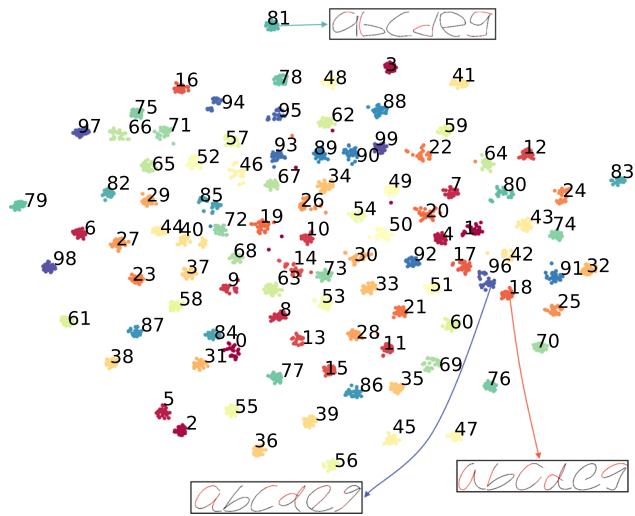

**Fig. 16** T-SNE Maaten, L.v.d., Hinton, G. (2008) visualization of the learned features before the FC layer and some writings of different clusters on LERID. The number of each cluster denotes the writer ID, and points of the same colour represent writings of the same writers. Most identities are distinguishable from each other

#### 4.4.6 Feature Embedding Visualization

To further analyse the learned feature embedding, we visualize the learned features by t-SNE (Maaten, L.v.d., Hinton, G. 2008). As shown in Fig. 16, the similar writings of different identities lie closely, e.g., the writings of writers "18" and "96", and the identities of the writers still can be distinguished from each other. Additionally, dissimilar writings are far from each other in the learned feature space, e.g., the writings of writers "18" and "81".

## 5 Conclusion

In this work, we address the *letter-level online writer-id* problem to make writer-id more practical, particularly on mobile devices. To address the large variance among online writing styles (Var-O-Styles) challenge in *letter-level online writer-id*, a novel deep framework is proposed, which includes a multi-branch encoder, letters and styles adapter (LSA) and hierarchical attention pooling (HAP). Additionally, a large LEtter-level online wRiter-ID (LERID) dataset is also contributed for more evaluation. By means of extensive experiments, we have shown that the components in the novel deep framework are beneficial for mining effective cues for discrimination, alleviating the effect of varying online writing styles, extracting letter-specific features, and finally aggregating available features step by step via attentions in multiple layers to achieve *letter-level online writer-id*. The proposed model is verified to be experimentally robust, even when using only a few letters and under the letter-independent setting.

**Acknowledgements** This work was supported partially by the National Key Research and Development Program of China (2018YFB1004903), NSFC(U1911401,U1811461), Guangdong Province Science and Technology Innovation Leading Talents (2016TX03X157), Guang-dong NSF Project (No. 2018B030312002), Guangzhou Research Project (201902010037), and Research Projects of Zhejiang Lab (No. 2019KD0AB03), and the Key-Area Research and Development Program of Guangzhou (202007030004). The corresponding author and principal investigator for this paper is Wei-Shi Zheng.# References

Arandjelović, R.,& Zisserman, A. (2012) Three things everyone should know to improve object retrieval. In *Computer vision and pattern recognition* (pp. 2911–2918). IEEE.

Bertolini, D., Oliveira, L. S., Justino, E., & Sabourin, R. (2013). Texture-based descriptors for writer identification and verification. *Expert Systems with Applications*, *40*, 2069–2080.

Bulacu, M., & Schomaker, L. (2007). Text-independent writer identification and verification using textural and allographic features. *IEEE Transactions on Pattern Analysis and Machine Intelligence*, *29*, 701–717.

Carlucci, F.M., Porzi, L., Caputo, B., Ricci, E., & Bulò, S.R. (2017) Just dial: Domain alignment layers for unsupervised domain adaptation. In *International conference on image analysis and processing* (pp. 357–369). Springer.

Chaabouni, A., Boubaker, H., Kherallah, M., Alimi, A.M., El Abed, H. (2011) Multi-fractal modeling for on-line text-independent writer identification. In *International conference on document analysis and recognition* (pp. 623–627). IEEE.

Chen, K. T. (1958). Integration of paths: A faithful representation of paths by noncommutative formal power series. *IEEE Transactions of the American Mathematical Society*, *89*, 395–407.

Chen, Z., Yu, H.X., Wu, A., & Zheng, W.S. (2018) Letter-level writer identification. In *Automatic face & gesture recognition* pp. 381–388. IEEE.

Christlein, V., Maier, A. (2018) Encoding CNN activations for writer recognition. In *International association for pattern recognition* (pp. 169–174). IEEE.

Dhingra, B., Liu, H., Yang, Z., Cohen, W.W., Salakhutdinov, R. (2017) Gated-attention readers for text comprehension. In *Meeting of the association for computational linguistics* (pp. 1832–1846). ACL.

Dwivedi, I., Gupta, S., Venugopal, V., & Sundaram, S. (2016) Online writer identification using sparse coding and histogram based descriptors. In *International conference on frontiers in handwriting recognition* (pp. 572–577). IEEE.

El Abed, H., Märgner, V., Kherallah, M.,& Alimi, A.M. (2009) Icdar 2009 online arabic handwriting recognition competition. In *International conference on document analysis and recognition* (pp. 1388–1392). IEEE.

Feng, M., Xiang, B., Glass, M.R., Wang, L., & Zhou, B. (2015) Applying deep learning to answer selection: a study and an open task. In *Automatic speech recognition and understanding* (pp. 813–820). IEEE.

Fu, J., Liu, J., Tian, H., Li, Y., Bao, Y., Fang, Z., & Lu, H. (2019) Dual attention network for scene segmentation. In *Computer vision and pattern recognition* (pp. 3146–3154). IEEE.

Gargouri, M., Kanoun, S.,& Ogier, J.M. (2013) Text-independent writer identification on online arabic handwriting. In *International conference on document analysis and recognition* (pp. 428–432). IEEE.

Hochreiter, S., & Schmidhuber, J. (1997). Long short-term memory. *Neural Computation*, *9*(8), 1735–1780.

IAM On-Line Handwriting Database. http://www.fki.inf.unibe.ch/databases.

Ioffe, S., & Szegedy, C. (2015) Batch normalization: Accelerating deep network training by reducing internal covariate shift. In *International conference on machine learning* (pp. 448–456). IEEE.

Khan, F. A., Khelifi, F., Tahir, M. A., & Bouridane, A. (2018). Dissimilarity gaussian mixture models for efficient offline handwritten text-independent identification using sift and rootsift descriptors. *IEEE Transactions on Information Forensics and Security*, *14*, 289–303.

Lai, S., & Jin, L. (2019) Offline writer identification based on the path signature feature. In *International conference on document analysis and recognition* (pp. 1137–1142). IEEE.

Lai, S., Jin, L., Lin, L., Zhu, Y., & Mao, H. (2020) Synsig2vec: Learning representations from synthetic dynamic signatures for real-world verification. In *Association for the advancement of artificial intelligence* (pp. 735–742). AAAI.

Li, B., Sun, Z., & Tan, T. (2007) Online text-independent writer identification based on stroke's probability distribution function. In *International conference on biometrics* (pp. 201–210). Springer.

Li, B., & Tan, T. (2009) Online text-independent writer identification based on temporal sequence and shape codes. In *International conference on document analysis and recognition* (pp. 931–935). IEEE.

Lin, Z., Feng, M., Santos, C. N. D., Yu, M., Xiang, B., Zhou, B.,& Bengio, Y. (2017) A structured self-attentive sentence embedding. In *International conference on learning representations*.

Liu, W., Wen, Y., Yu, Z., Li, M., Raj, B., & Song, L. (2017) Sphereface: Deep hypersphere embedding for face recognition. In *Computer vision and pattern recognition* (pp. 212–220). IEEE.

Maaten, L.v.d., Hinton, G., (2008). Visualizing data using t-SNE. *Journal of Machine Learning Research*, *9*, 2579–2605.

Moon, H., & Phillips, P. J. (2001). Computational and performance aspects of pca-based face-recognition algorithms. *Perception*, *30*, 303–321.

Nam, H., Ha, J.W., & Kim, J. (2017) Dual attention networks for multimodal reasoning and matching. In *Computer vision and pattern recognition* (pp. 299–307). IEEE.

Namboodiri, A., & Gupta, S. (2006) Text independent writer identification from online handwriting. In *International conference on frontiers in handwriting recognition* (pp. 566–571).

Nasuno, R., Arai, S. (2017) Writer identification for offline japanese handwritten character using convolutional neural network. In *International conference on intelligent systems and image processing* (pp. 94–97). Springer.

Nguyen, H. T., Nguyen, C. T., Ino, T., Indurkhya, B., & Nakagawa, M. (2019). Text-independent writer identification using convolutional neural network. *Pattern Recognition*, *121*, 104–112.

Ramaiah, C., Shivram, A., & Govindaraju, V. (2013) A bayesian framework for modeling accents in handwriting. In *International conference on document analysis and recognition* (pp. 917–921). IEEE.

Sae-Bae, N., & Memon, N. (2014). Online signature verification on mobile devices. *IEEE Transactions on Information Forensics and Security*, *9*, 933–947.

Schlapbach, A., & Bunke, H. (2007) Fusing asynchronous feature streams for on-line writer identification. In *International conference on document analysis and recognition* (Vol. 1, pp. 103–107). IEEE.

Schlapbach, A., Liwicki, M., & Bunke, H. (2008). A writer identification system for on-line whiteboard data. *Pattern Recognition*, *41*(7), 2381–2397.

Schuster, M., & Paliwal, K. K. (1997). Bidirectional recurrent neural networks. *Signal Processing*, *45*, 2673–2681.

Shivram, A., Ramaiah, C., & Govindaraju, V. (2013). A hierarchical bayesian approach to online writer identification. *Iet Biometrics*, *2*(4), 191–198.

Si, J., Zhang, H., Li, C.G., Kuen, J., Kong, X., Kot, A.C., & Wang, G. (2018) Dual attention matching network for context-aware feature sequence based person re-identification. In *Computer vision and pattern recognition* (pp. 5363–5372). IEEE.